\documentclass[letterpaper]{article}
\usepackage{aaai21}
\usepackage{times}
\usepackage{helvet}
\usepackage{courier}
\usepackage{algorithm}
\usepackage[noend]{algpseudocode}
\usepackage{graphicx}
\usepackage{xcolor}
\usepackage[normalem]{ulem}
\usepackage{amsmath}
\usepackage{mathtools}
\usepackage{amsfonts}

\usepackage{natbib}

\begin{document}

\title{ReMix Training for Calibrated Imbalanced Deep Learning}

\author{Colin Bellinger\textsuperscript{\rm 1}, 
        Roberto Corizzo\textsuperscript{\rm 2}, 
        Nathalie Japkowicz\textsuperscript{\rm 2} 
        \\
}

\affiliations{

    \textsuperscript{\rm 1} National Research Council of Canada \\ Ottawa, Canada\\
    \textsuperscript{\rm 2} American University \\ Department of Computer Science \\ Washington, DC 20016, USA\\
    colin.bellinger@nrc-cnrc.gc.ca,rcorizzo@american.edu,japkowic@american.edu
}
\maketitle
\begin{abstract}
\begin{quote}
Class imbalance is a problem of significant importance in applied deep learning where trained models are exploited for decision support and automated decisions in critical areas such as health and medicine, transportation, and finance. The challenge of learning deep models from imbalanced training data remains high, and the state-of-the-art solutions are typically data dependent and primarily focused on image data. Real-world imbalanced classification problems, however, are much more diverse thus necessitating a general solution that can be applied to tabular, image and text data. In this paper, we propose ReMix, a training technique that leverages batch resampling, instance mixing and soft-labels to enable the induction of robust deep models for imbalanced learning. Our results show that dense nets and CNNs trained with ReMix generally outperform the alternatives according to the g-mean and are better calibrated according to the balanced Brier score.
\end{quote}
\end{abstract}

\section{Introduction}\label{sec:intro}

There is a growing amount of interest in applying deep learning to complex and critical domains, such has medicine, health and safety and finance \cite{rao2006data,sanz2014compact}, that exhibit both imbalanced class priors and asynchronous misclassification costs.  To be safely applied the models must achieve high recall on the minority classes, and be well-calibrated. Deep learning algorithms, however, have been shown to exhibit unsatisfactory predictive performance on poorly represented classes \cite{buda2018systematic,cao2019learning}. Moreover, they have recently been demonstrated to be poorly calibrated \cite{guo2017calibration} and drastically shift their prediction with small changes in the input space \cite{devries2018learning} leading to safety concerns related to the real-world application of AI \cite{amodei2016concrete}. These issue are exacerbated by training sets with imbalanced priors and have significantly limited the potential to apply the power of deep learning to many critical real-world domains.

\begin{figure}
\centering
    \includegraphics[scale=0.22]{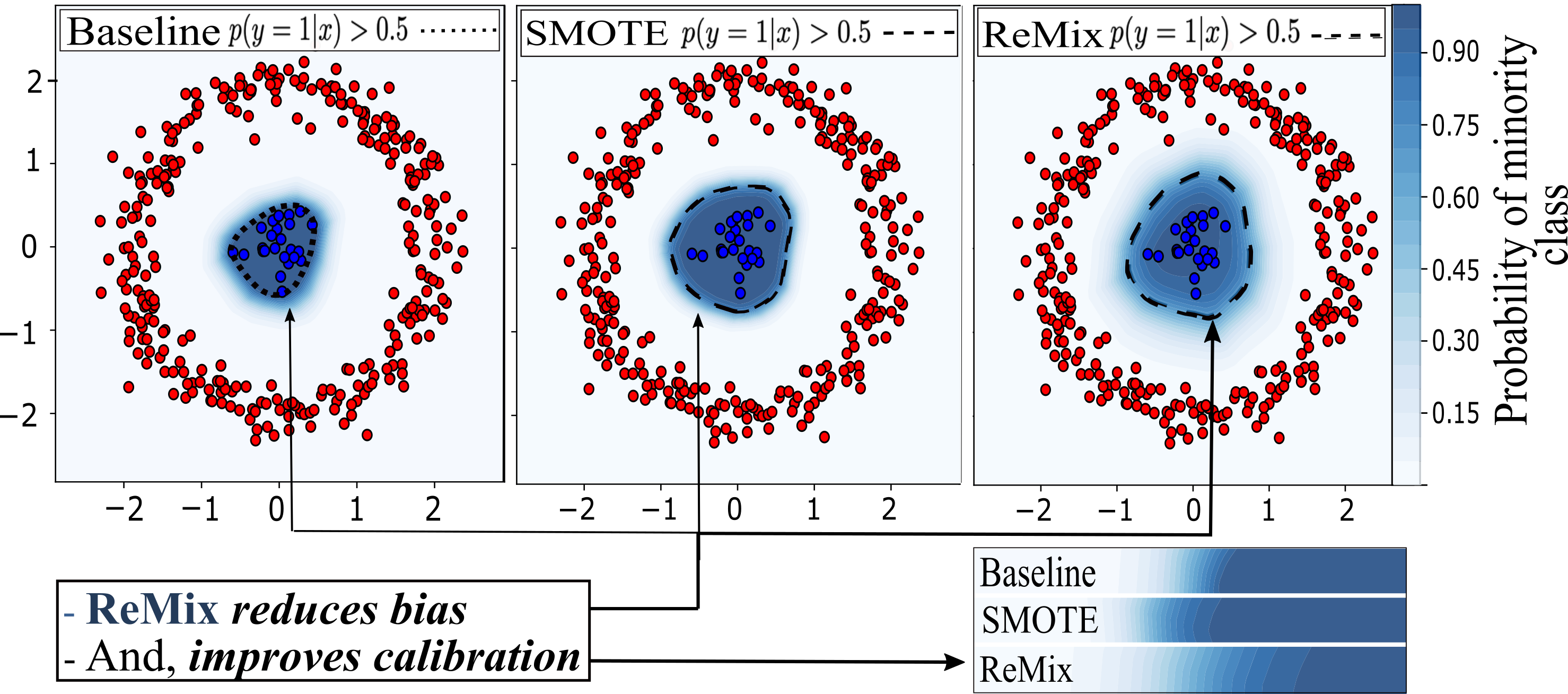}
\caption{Demonstration of neural network decision surfaces and classification boundaries ($p(y=1 |x) > 0.5$) for the ring dataset. Network trained on imbalanced data (left), the network training data balanced via SMOTE (centre), and the network trained on ReMix mini-batches (right). The decision boundary and gradient of minority class probability demonstrates that ReMix reduces the bias and improves calibration.}
\label{fig:remixDemo}
\end{figure}

Traditional methods to deal with class imbalance involve resampling (random undersampling the majority class, random oversampling the minority class and generating additional synthetic minority samples) or cost-adjustment \cite{branco2016survey}. Synthetic resampling methods based on SMOTE \cite{chawla2002smote} are generally preferred as they are simple to apply preprocessing steps that are classifier independent. SMOTE-based methods generate synthetic samples by random interpolation between \textit{k}-nearest neighbours in the minority class. Whilst these methods have been shown to improve the predictive performance of shallow models, they do not improve calibration with respect to underrepresented classes \cite{wallace2012class}. 

\begin{figure*}
\centering
    \includegraphics[scale=0.26]{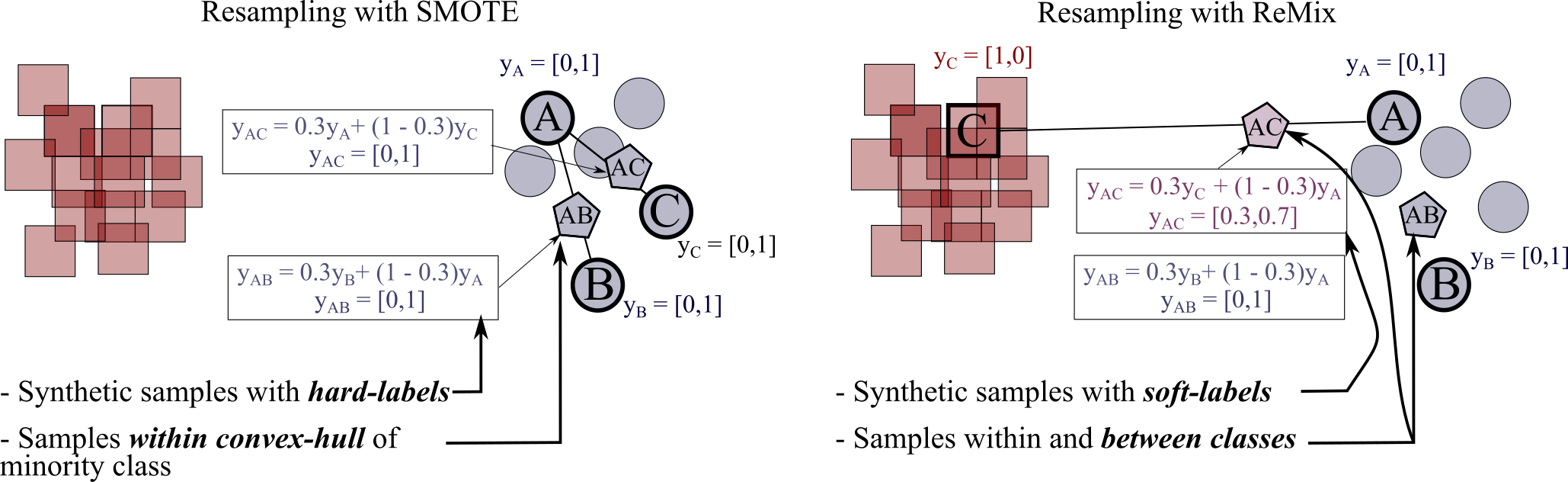}
\caption{Demonstration of samples generated by by SMOTE versus ReMix.}
\label{fig:SmoteVsRemix}
\end{figure*}

In the context of deep learning, it is generally not possible to load and preprocess the entire dataset. Alternatively, the resampling is applied within each mini-batch \cite{huang2016learning,buda2018systematic}. Calculating \textit{k}-nearest neighbours for synthetic oversampling in each mini-batch slows training and can be intractable for large-scale problems. In addition, the synthetic distribution generated by SMOTE has been shown to contract the minority class distribution \cite{elreedy2019comprehensive}. This places an upper-bound on the potential reduction in predictive bias, which is exacerbated in the context of mini-batches where only a subset of minority samples may appear. In addition, SMOTE has been shown to negatively alter the class distribution when factors such as disjoint data distributions, noise, and outliers are present \cite{bellinger2019framework}. 

Recently, proposed strategies to handle class imbalance in deep learning have focused on the generation of additional synthetic samples via GANs and VAEs to balances the training set \cite{mullick2019generative,wang2019wgan,dai2019generative}. These methods typically focus in image classification and necessitate the learning of additional models and/or significantly more parameters. These learning processes are computationally costly and are themselves susceptible to imbalanced classes and rare training instances. As a result, solutions of this nature may only improve predictive performance on applications with little to moderate imbalance. Moreover, these method do not improve calibration because they generate with-in manifold samples and are assigned hard labels for training. 

To improve performance on imbalanced deep learning problems, we propose the ReMix training algorithm. ReMix adds minimal computation and algorithmic complexity and achieves a significant performance advantage on a wide cross-section of imbalance domains. It is an advancement of MixUp training \cite{zhang2017mixup} that is optimized for imbalanced learning. Whereas SMOTE applies random undersampling to the majority classes and synthetic oversampling to the minority classes, ReMix efficiently samples approximately balanced mini-batches of synthetic training data with soft-labels from the vicinity distribution $\nu(\cdot)$ of the real data. 

Repeatedly sampling balanced mini-batches from $\nu(\cdot)$ has the following advantages over SMOTE-based methods: \textit{a)} It reduces the risk of overfitting because the likelihood of sampling a feature vector multiple times is low. \textit{b)} It avoids the risk of information loss due to randomly undersampling the majority class. \textit{c)} It reduce prediction bias due to the smooth spread of the sampled data points beyond the convex-hull of the minority class data.

The use of soft-labels has been shown to regularize deep neural networks (NN) and promote linear behavior in-between training examples \cite{zhang2017mixup}; in addition, they improve calibration in NN \cite{guo2017calibration}. We postulate that in addition to the previously discussed benefits, soft-labels in the context of resampling for imbalanced learning enables the safe generation of minority samples beyond the convex-hull of the minority class, and reduces the risk that noisy samples will have a strong impact on the induced model.

Figure \ref{fig:remixDemo} presents a comparison of a baseline NN model trained on imbalanced data (left), a model trained with SMOTE (centre), and ReMix training (right). Each plot shows the corresponding decision surface (blue gradient) and soft-max decision boundary (dashed line). The baseline model has a significant predictive bias in favour of the majority class, whereas the models trained with SMOTE and ReMix have decision boundaries pushed out from the minority samples. This indicates that both methods reduce the predictive bias, with ReMix having slightly greater impact.

In addition to reducing the bias, this figure demonstrates that the NN trained with ReMix learns to have a smooth, linear decrease in $p_{\theta_{ReMix}}(y=1|x)$ between classes. In practice, this behaviour between classes is associated with better calibrations. The change in predicted class probably is depicted by the shading from dark blue (high probability of minority class) to light blue (low probability of minority class). The baseline and SMOTE trained models both exhibit a sharp shift in probability at their decision boundaries, which is associated with poor calibration.

Our empirical results on benchmark tabular and image datasets show that deep learning models training on imbalanced datasets using ReMix, achieve equivalent or better predictive performance, and better calibration than models trained with traditional resampling and cost-adjustment methods.

\begin{figure*}
\centering
    \includegraphics[scale=0.26]{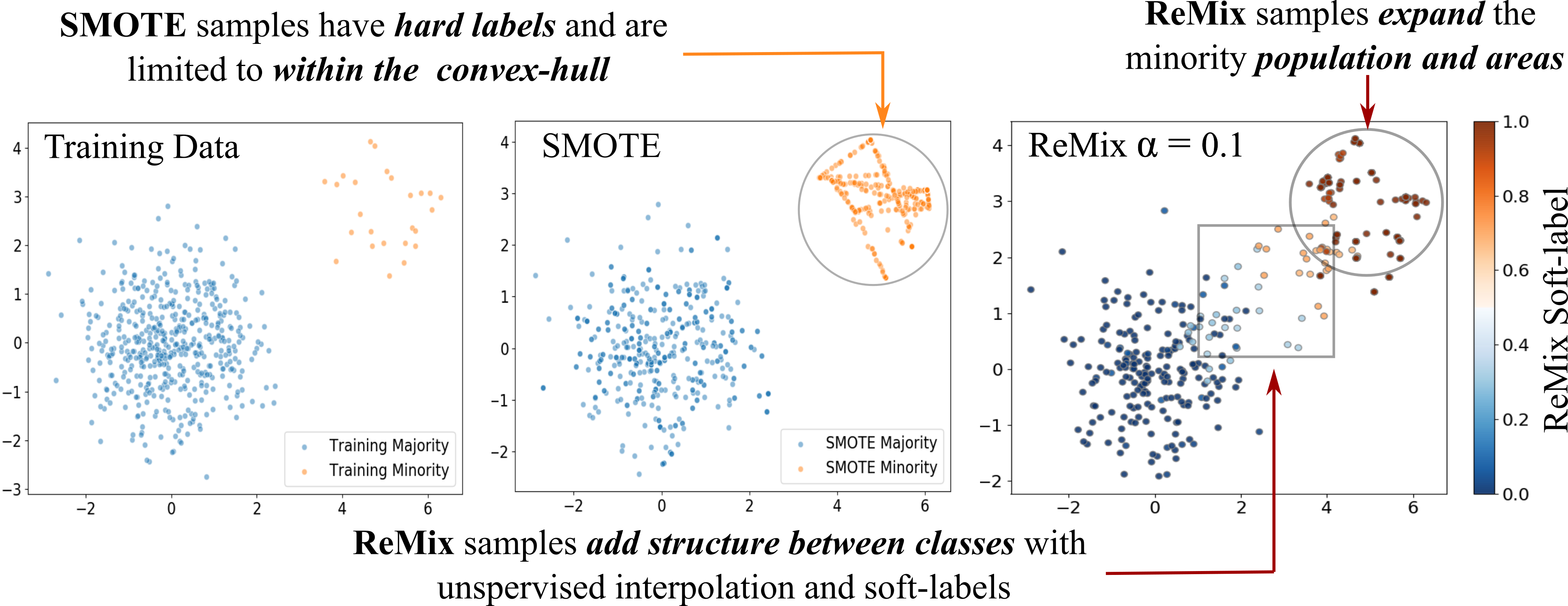}
    \vspace{0.1cm}
\caption{Comparison of samples generated with ReMix and SMOTE.}
\label{fig:remixSampleCom}
\end{figure*}

\section{Related Work}\label{sec:relatedWork}

The authors in \cite{anand1993improved}, studied the impact of training NNs on imbalanced classification data and found that the majority class errors dominate the gradients-based weight updates during training. Cost-adjustment or re-sampling the examples in the mini-batch are standard techniques to deal with the predictive bias in the resulting NN model \cite{huang2016learning,buda2018systematic,cui2019class,johnson2019survey}. Unlike less powerful shallow models, when resampling or cost-adjustment are applied during the training of highly parameterized deep models, they cause the models to over-fit the limited information in the minority classes \cite{cao2019learning}. Contrary to existing methods, ReMix increases the frequency and diversity of the minority training samples in successive mini-batches by sampling from $\nu(\cdot)$ thereby reducing the predictive bias and the risk of over-fitting.

It is increasingly being recognized that decision-making systems must be both accurate and calibrated \cite{wallace2012class,guo2017calibration,thulasidasan2019mixup,huang2020experimental}. In \cite{wallace2012class}, the authors examined the calibration of shallow models trained on imbalanced data and found that the resulting minority class predictions were poorly calibrated. In addition, they found that standard methods of mitigating the imbalance do not improve model calibration. Whilst, modern deep NNs have recently been demonstrated to be poorly calibrated \cite{niculescu2005obtaining}, the authors in \cite{guo2017calibration} found that the combined effect of feature mixing and soft-labels significantly improves the calibration in deep models in balanced class settings. Moreover, they found that MixUp is superior to other strategies for improving calibration in deep models. Similarly, the prioritized feature mixing and soft-labels in ReMix enables it to both achieve a predictive performance that is superior to standard resampling methods, such as SMOTE, and improved calibration. 

\begin{figure*}
\centering
    \includegraphics[scale=0.3]{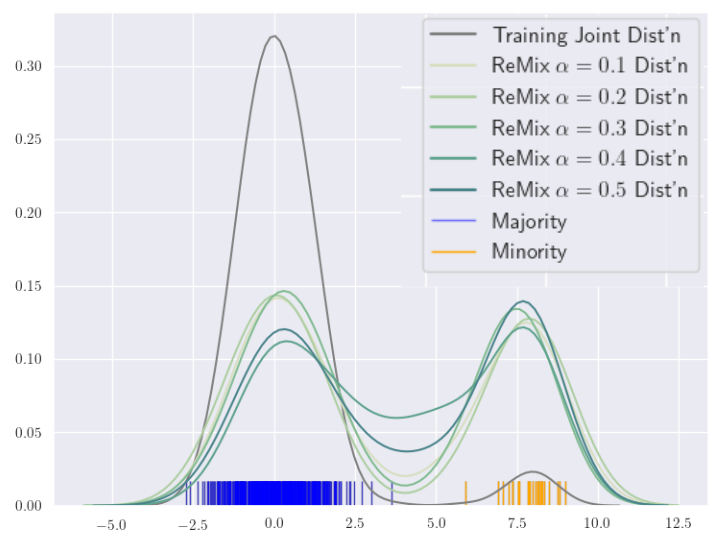}
    \hspace{0.5cm}  
    \includegraphics[scale=0.3]{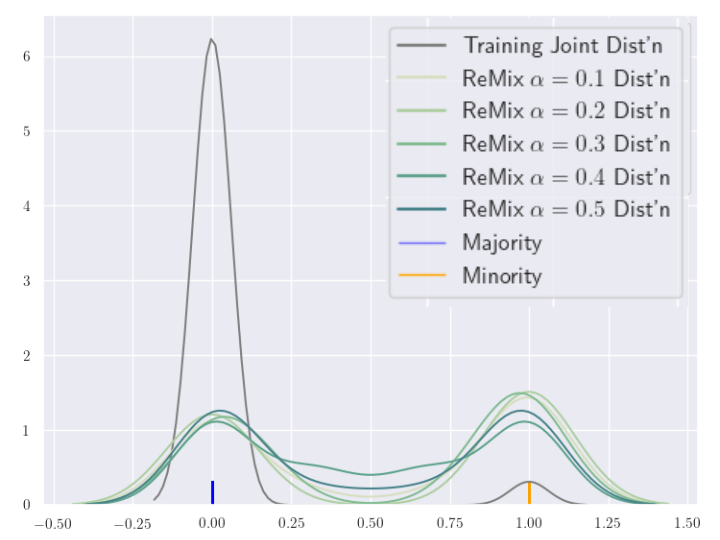}

\caption{Kernel estimated data (left) and label (right) distributions for ReMix with respect to $\alpha$-values.}
\label{fig:remixDataAndLabelDistn}
\end{figure*}

\section{ReMix: Resampling MixUp}\label{sec:mixupResampling}

\subsection{MixUp}

MixUp is based on the principle of Vicinal Risk Minimization (VRM) \cite{chapelle2001vicinal}. Under this principle, the model is trained on mini-batches $\mathcal{B}_\nu := \{(\tilde{x}_i, \tilde{y}_i)\}^m_{i=1}$ of synthetic feature-target pairs $(\tilde{x}, \tilde{y})$ drawn from the vicinity, $\nu(\cdot)$, of the true training data $X$ and labels $Y$. The empirical vicinal risk of a model $f$ is minimized over $\mathcal{B}_\nu$ as:
\begin{equation}
    R_\nu(f) = \frac{1}{m}\sum^m_{i=1} l(f(\tilde{x}_i),\tilde{y}_i).
\end{equation}
MixUp utilizes the generic vicinal distribution:
\begin{multline}
    \mu(\tilde{x}_i, \tilde{y}_i | x_i, y_i) = \frac{1}{n} \sum^n_{j=1} \mathop{E}_{\lambda} [\delta(\tilde{x}= \lambda \tilde{x}_i + (1-\lambda) \tilde{x}_j,\\
    \tilde{y} = \lambda \tilde{y}_i + (1-\lambda)\tilde{y}_j)].
\end{multline}
In order to approximate vicinal sampling in an efficient and domain agnostic way, MixUp creates convex combinations of random training pairs $x_i$ $x_j$ and their one-hot labels $y_i$ $y_j$:
\begin{align*}
    \tilde{x} &= \lambda x_i + (1-\lambda)x_j\\
    \tilde{y} &= \lambda y_i + (1-\lambda)y_j,
\end{align*}
where $\lambda \in [0,1]$ is independently sampled from a Beta distribution $Beta(\alpha, \alpha)$ with $\alpha \in [0,\infty]$ for each mini-batch of training. The $\alpha$-value is a user-specified as a hyper-parameter that controls the amount of mixing between training samples.

Mixing is performed in an unsupervised manner, which enables the mixing of samples from heterogeneous classes. Because mixing is performed by \textit{i.i.d.} sampling from the data, the prior class probability of is approximately maintained in $\mathcal{B}_\nu$. As a result, when applied to imbalanced data, MixUp produces a model that is biased, poorly calibrated and fit with respect to the minority class.

\subsection{ReMix Algorithm}

ReMix is a training method for binary and multi-class imbalanced deep learning.  Unlike existing methods, it serves to improve accuracy and calibration on the poorly represented classes. 

ReMix is presented in Algorithm \ref{alg:remix}: during training, it is iteratively called to sample mini-batches of synthetic training data from $\nu(\cdot)$. Prioritized sampling with replacement is used to produce approximately equally class represented in each mini-batch. This is implemented as a three-step process in which \textit{a}), a mini-batch $\mathcal{B}_i$ is sampled from the training data in the standard way, \textit{b}) weighted resampling is applied to $\mathcal{B}_i$ produce a balanced set $\mathcal{B}_i^\prime$, and then \textit{c}) the final mini-batch $\mathcal{B}_i^{\prime\prime}$ is formed by convexly combining a random subset of instances and one-hot labels from $\mathcal{B}_i^\prime$.

\begin{algorithm}[H]
\caption{ReMix Algorithm.}
\label{alg:remix}
\textbf{Input:} Beta parameter $\alpha \ge 0$, mini-batch size $B$\\
\textbf{Output:} Balanced random mini-batch $X^\prime$ sampled from $\nu(X,y)$\\
\textbf{Algorithm:}
\begin{algorithmic}[1]
\State{Sample the next mini-batch of features $X$ and one-hot labels $Y$}
\State{Set $C$ equal to the number of classes in $Y$}
\State{Sample $B/C$ instances from each class into $X', Y' = \text{resample}(X, Y, B/C)$}
\State{Generate a random permutation $\text{idx} = \text{permutation}(b)$}  
\State{Sample a mixing parameter $\lambda = \text{Beta}(\alpha, \alpha)$} 
\State{MixUp features $X'' = \lambda \times X' + (1.0 - \lambda) \times X^\prime[\text{idx}]$}
\State{MixUp labels $Y'' = \lambda \times Y' + (1.0 - \lambda) \times Y^\prime[\text{idx}]$}\\
\Return{ReMixed mini-batch $X''$, $Y''$}    
\end{algorithmic}
\end{algorithm}

\subsection{Demonstration}

Figure \ref{fig:SmoteVsRemix} contrasts the synthetic samples generated by SMOTE with those produced by ReMix. SMOTE generates synthetic samples with hard labels within the convex-hull formed by the minority class. This leads to overfitting the resampled convex-hull and fails to address poor calibration. Alternatively, ReMix samples are smoothly spread over the joint distribution of the classes. This leads to a reduction in predictive bias, better generalization and calibration.  

Figure \ref{fig:remixSampleCom} illustrates the application of SMOTE and ReMix to a 2-dimensional, imbalanced, binary datast. It illustrates mini-batch data produced by ReMix and SMOTE. The figure shows that ReMix expands the minority space, helping to reduce the prediction bias of the induced model. In addition, it demonstrates how mixing with soft-labels add structure between the classes. The corresponding impacts on the learned models were highlighted in Figure \ref{fig:remixDemo} in the Introduction.

Figure \ref{fig:remixDataAndLabelDistn} illustrates how the user-specified $\alpha$-value shapes the feature and soft-label distribution in ReMix. The first two plots of the figure show the kernel density estimation of the synthetic samples and soft-labels. The grey line shows the distribution of the original 1-dimensional Gaussian training set. The corresponding majority class samples (blue) and minority class samples (orange) are shown on the bottom of each plot. The green lines show the distribution of the ReMix data for $\alpha$ ranging from 0.1 to 0.5 (light to dark). The plots show that higher $\alpha$ values produce more mixing of the features vectors and labels. This adds more structure between the classes, however, setting $\alpha > 0.5$ can cause a degradation in performance.

\section{Experimental Setup}\label{sec:setup}

\begin{table*}
    \centering
    \caption{Details of the binary and multi-class datasets used in the following experiments. Dataset from source A are available at \cite{uci2019}, and those from source B are available at \cite{lecun1998}.}
    \begin{tabular}{lccccc}
        \hline
         Dataset &  Source & Classes & Dim & Inst & IRs\\
         \hline
         \hline
         Musk & A  & 0 vs 1 & 168 & 6,597 & 0.01, 0.025, 0.05  \\
         Segment & A  & 1 vs $2..7$ & 19 & 2,310 & 0.01, 0.025, 0.05 \\
         Statlog & A  & 1 vs $2..7$ & 36 & 6,435 & 0.01, 0.025, 0.05 \\
         Seizure & A  & 1 vs $2..5$ & 179 & 11,500 & 0.05, 0.025, 0.01 \\
         coil2000 & A  & 0 vs 1 & 86 & 9,000 & 0.05, 0.025, 0.01 \\
         Ozone & A  & 0 vs 1 & 73 & 2,536 & 0.025, 0.01 \\
         APS & A  & -1 vs 1 & 171 & 60,000 & 0.01 \\
         CIFAR 10 & B  & dog vs Other & $32 \times 32$ & 50,000 & 0.25, 0.1, 0.05 \\
         \hline
         \hline
         Seizure & A  & $1,2$ vs $3,4,5$ & 179 & 11,500 & 0.25, 0.1, 0.05 \\
         Optical digits & A  & $1,2,3$ vs $0,4...9$ & 64 & 5,620 & 0.25, 0.1, 0.05 \\
         landsatSatellite & A  & $2,3,4$ vs $1,5,6,7$ & 36 & 6,435 & 0.25, 0.1, 0.05 \\
         CIFAR 10 & B  & automobile,bird,cat vs Other & $32 \times 32$ & 50,000 &  0.25, 0.1, 0.05 \\
         \hline
    \end{tabular}
    \label{tab:datasets}

\end{table*}

\subsection{Datasets}

The classification experiment are conducted on 1 image dataset and 7 tabular datasets. We evaluate both binary and multi-class setups. For the purpose of these experiments, the datasets were converted to imbalanced classification problems. Multi-class imbalanced datasets were formed by selecting a subset of classes to randomly downsampled, and binary imbalanced datasets were created by downsampling a signle class and merging the rest. 

Class imbalance ratios ($IR = \frac{N^-}{N^+}$) in the range of 0.1 to 0.01 were created. For each dataset, the lower limited was determined by the original class sizes. The specifics of the datasets are outlined in Table \ref{tab:datasets}.

\subsection{Deep Learning Models}

In the tabular data experiments, we utilize a three layer deep NN with 0.1 dropout and relu activation. For the image data, we employ a convolutional NN with relu activation, two $3\times 3$ convolutional layers and two $2\times 2$ max pooling layers interleaved, and a dense layer. Both networks architectures utilize  categorical cross-entropy loss with ADAM optimization. These represent common baseline architectures for the target datasets. 

Whilst optimizing the architecture and training of the networks for individual datasets is likely to produce slight improvements in performance, the focus of this work is on the relative effectiveness of the rebalanced methods, which is unlikely to be affected by further optimization. Furthermore, we utilized moderate sized deep nets and datasets for the purpose of these experiments as they reduce energy consumption and time, whilst effectively testing our hypothesis.

\subsection{Resampling Methods}

The following experiments are preformed with above described deep NN models train on the imbalanced data (baseline). This is compared to training with MixUp, cost adjustment, with SMOTE resampling in the mini-batch and ReMix. 

\subsubsection{Metrics and Evaluation}

We evaluate the classification performance in terms of g-mean (GM) and the balanced brier score (BBS). The GM is a standard metric for evaluating imbalanced classifiers. It calculates the geometric mean of the true positive rate and true negative rate \cite{buda2018systematic}.

The Brier score (BS) assess the calibration of the classifier. The score ranges from 0 to 1 with lower values indicating better calibration. Much like accuracy, however, is an imperfect metric for imbalance classification as it is biased toward performance on the majority class. Based on this observation, \cite{wallace2012class} proposed the use of class specific Brier scores for binary classification problems: 
\begin{align*}
    BS^+ &= \frac{\sum_{y_i=1} (y_i - \hat{P}(y_i | x_i)^2)}{N_+}\\
    BS^- &= \frac{\sum_{y_i=0} (y_i - \hat{P}(y_i | x_i)^2)}{N_-}.
\end{align*}

To simplify the calibration analysis, we propose the arithmetic mean of the Brier score calculated independently for each class (Balanced Brier score). 
\begin{equation}
    BBS = \frac{\sum_j^K BS_j}{K}.
\end{equation}
This treats the calibration of each class as equally important, and is the calibration equivalent of the balanced per-class accuracy, which is a widely applied evaluation metric in imbalanced classification.

The results for each setup are recorded as the mean and standard deviation of calculated across $10 \times 2$-fold cross validation. During model learning $30$ percent of the training set is partitioned off for model validation. All experiments were preformed with Tensorflow 2 on Ubuntu 18.04 desktop running a GeForce RTX 2080 Ti GPU. 

\begin{table}
    \centering
    \caption{The sum of ranks on the binary and multi-class tabular datasets.}
    \begin{tabular}{l|cc|cc}
    \hline
    & \multicolumn{2}{c|}{Binary} &\multicolumn{2}{c}{Multi-class} \\
	& GM	& BBS & GM & BBS \\
	\hline
	\hline
    Baseline      &	65 &	63& 31	& 38 \\
    Cost Adjusted &	70 &	66& 28	& 36 \\
    SMOTE         &	36 &	54& 24	& 30 \\
    MixUp         &	77 &	63& 40	& 22 \\
    ReMix         &	22 &	22& 12	& 9 \\
    \hline
    \end{tabular}
    \label{tab:tabularDataAllBcMc}
\end{table}

\section{Results}\label{sec:results}

\begin{figure*}[h]
\centering
    \includegraphics[scale=0.23]{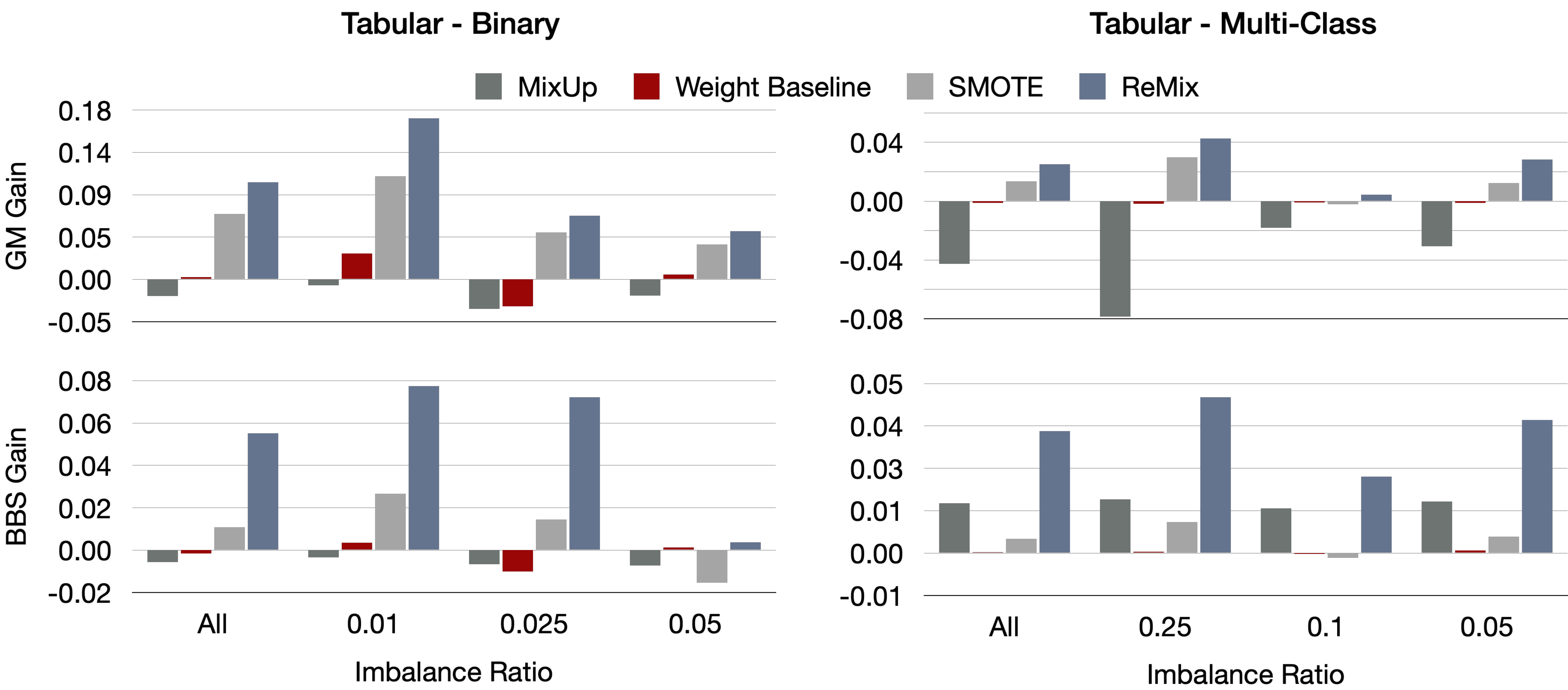}
\caption{Mean performance gains over the baseline on the tabular binary and multi-class classification data at different imbalance ratios.}
\label{fig:tabularGains}
\end{figure*}

\subsection{Tabular Datasets}

Table \ref{tab:tabularDataAllBcMc} summarizes the rank of the performance of each training strategy on the tabular datasets across all IR\footnote{Individual results for dataset including means and standard deviation are included in the supplementary material.}. Each cell shows the sum of ranks for the method. Rank 1 is the best rank, therefore, a lower sum of ranks indicates better performance in general. ReMix produces a much lower GM and BBS rank than the alternatives on both the binary and multi-class data. This indicates that it is generally a top performing method. 

Figure \ref{fig:tabularGains} provides further insight into the performance advantage of ReMix over the alternative techniques. It shows the mean performance gain over the baseline deep NN achieved by ReMix and the comparison methods. The left column reports results for the imbalance binary classification data and the right column shows the results for the imbalanced multi-class data. The top row includes the GM gain ($GM(f_{alternative}) - GM(f_{baseline})$), and the bottom row reports BBS gains ($GM(f_{baseline})) - GM(f_{alternative}$). A large positive score indicates a greater improvement over the baseline deep NN. Each plot shows the gain for IRs $0.05$, $0.025$ and $0.01$, along with the mean gain overall IRs (All). 

The results show that ReMix produces a greater mean improvement than the alternative methods. The greatest improvements in GM are produces by ReMix on the binary data. Both ReMix and SMOTE (the second best method in terms of GM), see their GM gains decline with more extreme imbalance. As hypothesized, the BBS Gain indicates that ReMix produces a much larger improvement in calibration than the alternative methods on the binary and multi-class tabular data.

\begin{figure*}[h]
\centering
    \includegraphics[scale=0.23]{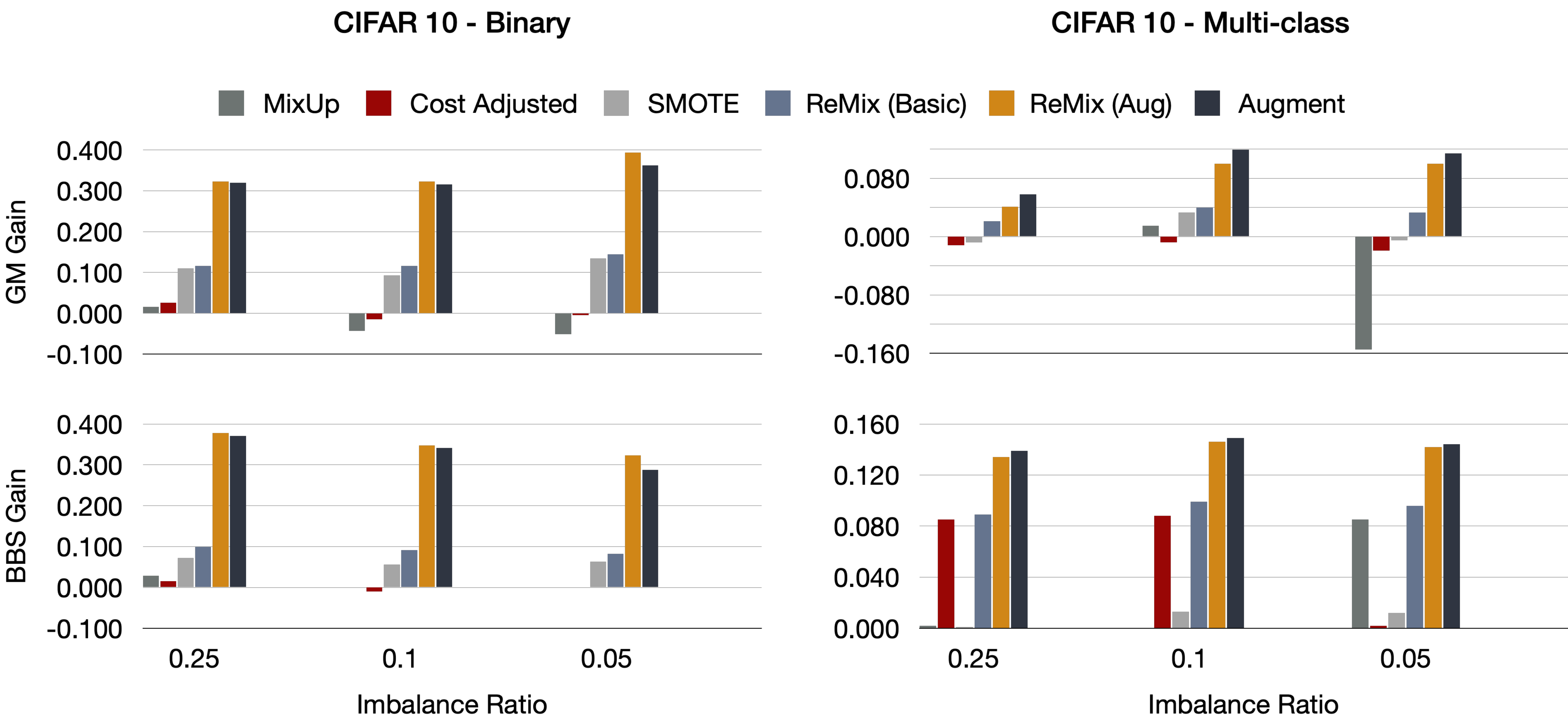}
\caption{Mean performance gains over the baseline on the tabular binary and multi-class classification data at different imbalance ratios.}
\label{fig:cifar10Gains}
\end{figure*}
 
\subsection{Image Datasets}

Image data provides a unique potential to apply domain specific data augmentations. These have been shown to improve performance \cite{krizhevsky2012imagenet}, and can be applied as a domain specific means of resampling to balance the training data. In this section, we explore the effectiveness of incorporating resampling image specific augmentations in ReMix. We list this as ReMix (Aug), and specify the standard version as ReMix (Basic). For completeness, we also include resampling with augmentation (Augment) without mixing.  We define augmentations in the range of 20 percent rotations, width shift range of 0.2 and height shift range of 0.2. 

Figure \ref{fig:cifar10Gains} shows the performance gain over the baseline deep NN achieved by ReMix and the comparison methods on the CIFAR 10 dataset\footnote{Full results including means and standard deviation are included in the supplementary material.}. The left column reports results for the imbalance binary classification version and the right column shows the results for the imbalanced multi-class version of the data. It shows that ReMix (Basic) outperforms MixUp, cost adjustment and SMOTE in terms of both GM gain and BBS gain. As before, however, the performance gains are smaller on the multi-class data. 

Resampling and ReMix with image specific augmentations shows a large improvement in GM gain and BBS gain over the alternative methods. On the binary data ReMix (Aug) produces the best results with Augment producing the second best gains. The order is switched on the multi-class data. This suggests that when domain specific augmentations are available and can be incorporated into resampling or ReMix, they can serve to further improve the performance and the calibration. 

\section{Discussion}\label{sec:dicussion}

\begin{figure*}
\centering
    \includegraphics[scale=0.27]{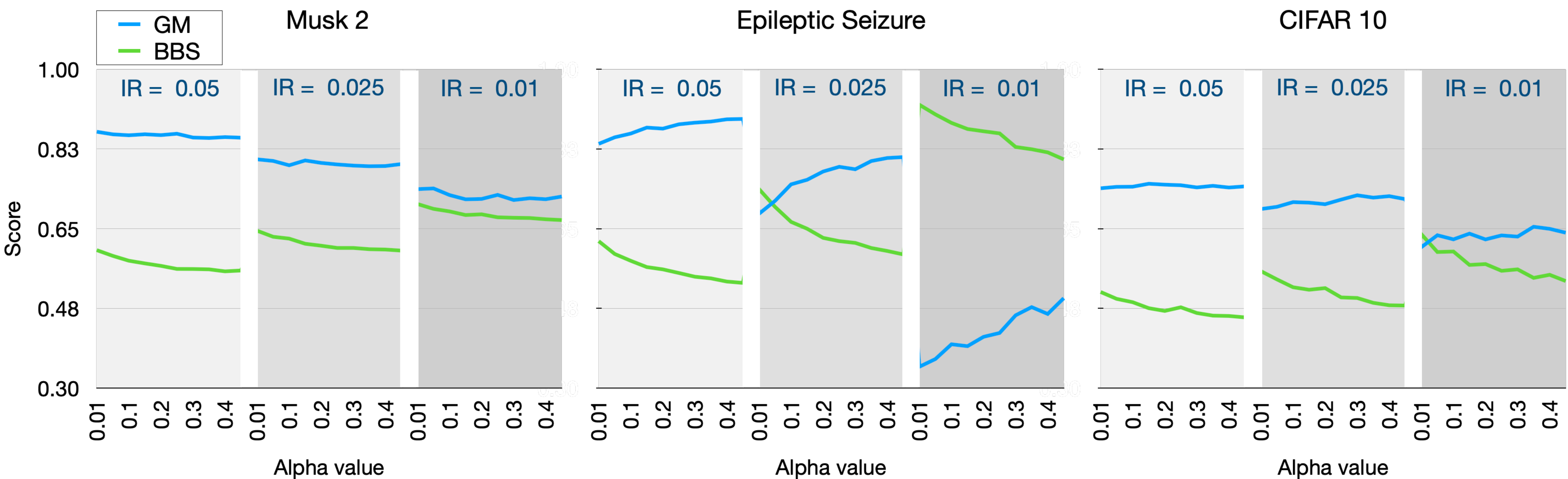}
\caption{Sensitivity of deep learning model to the $\alpha$ parameter in ReMix.}
\label{fig:remixParamSens}
\end{figure*}

\subsection{Sensitivity Analysis}

The ReMix results presented in the previous section utilize $\alpha=0.1$. Additional sensitivity analysis is required to understand the impact of the choice of $\alpha$ in the context of imbalance classification. Figure \ref{fig:remixParamSens} plots GM and BBS as a function of $\alpha$ on the Musk 2, Epileptic Seizure and CIFAR 10 datasets. 

Increasing values of $\alpha$ show lower BBS, which indicates better calibration. This results from the fact that large $\alpha$ values produce more structure between samples via feature and label mixing. The impact of $\alpha$ on GM is more dependent on the specific dataset. In the case of CIFAR 10, $\alpha$ in the range of $[0.01,0.4]$ has a negligible impact on GM. larger $\alpha$ values produce relatively large improvement in the metrics on the Epileptic Seizure dataset. This is particularly pronounced on the most extreme IR. Finally, larger $\alpha$ values causes subtle decrease in GM on the Musk 2 dataset. In general, GM appears to be stable for $\alpha$ between $0.01$ and $0.4$ and BBS decrease with $\alpha$ closer $0.4$. Thus, $\alpha=0.3$ is likely a good setting for ReMix. 

\subsection{Multi-class ReMix}

\begin{figure}
\centering
    \includegraphics[scale=0.25]{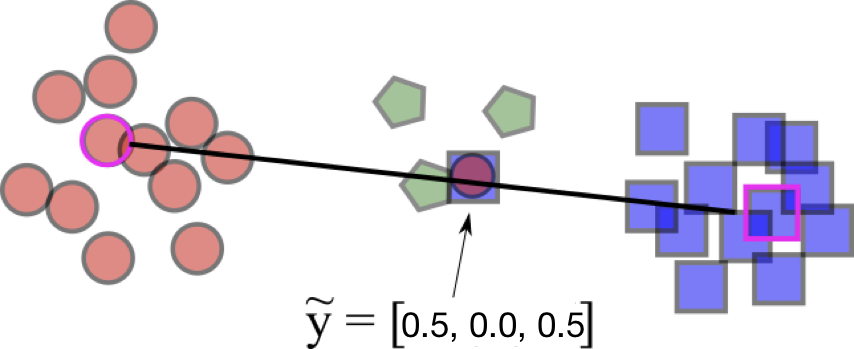}
\caption{Illustration of harmful interpolation between dispersed samples of opposing classes.}
\label{fig:remixErrMc}
\end{figure}

Whilst ReMix outperformed traditional imbalanced learning techniques on the binary and multi-class datasets, its advantage over the other methods declined in the multi-class setting. We hypothesize that in multi-class domains, mixing via unsupervised convex combinations of feature vectors can result in noisy, and possibly misleading samples. As demonstrated in Figure \ref{fig:remixErrMc}, this problem cannot be fully remedied by soft-labels in multi-class settings. The figure depicts a three class problem in which the synthetic sample contradicts the underlying data distribution. These samples may add variance to the mini-batch data and impact performance capabilities of the induced model. 

This problem can be addressed by setting a small $\alpha$-value to ensure the mixing parameter $\lambda$ sampled from the Beta distribution is near zero or one. This ensures a high probability of generating the sample close to its seeds (red circle or blue square).  This is not an entirely satisfactory approach, however, as setting $\alpha$ too low is likely to limit calibration. We leave further exploration of this question to future work.

\subsection{Image Classification}

ReMix outperformed the alternative methods on the CIFAR 10 data. However, the best results were produced with image specific data augmentations (ReMix (Aug) and Augment). This result highlights a potential limitation of feature level mixing and the data-specific augmentations for improving performance and calibration in imbalanced deep learning. 

In many applications and on many data types, domain specific data augmentations are not available or are too computationally expensive to integrate into deep NN training. On the one hand, our results show that improvements can be produced with standard ReMix trianing. On the other hand, the image classification results suggest that further research into the development and integration of general augmentation into ReMix could be very beneficial. If they can be made computationally efficient, domain-independent augmentation techniques, such as manifold oversampling \cite{bellinger2018manifold}, and GANs \cite{lim2018doping}, may prove to be effective. 

In addition, the incorporation of recent methods such as manifold MixUp \cite{verma2019manifold} and out-of-manifold data augmentation \cite{guo2020nonlinear}, into ReMix may serve to improve performance on multi-class and image data. 

\section{Conclusion}

Deep learning algorithms are increasingly being applied to critical domains involving class imbalance. In order for the models to be safe and effective tools, they must have both excellent predictive performance and be well calibrated. Nonetheless, the literature on imbalanced deep learning remains limited in scope and it is typically focused on optimized solutions for individual datasets and domains. 

In this work, we propose ReMix training for imbalanced deep learning. ReMix leverages batch resampling, instance mixing and soft-labels to enable the induction of robust deep models for imbalanced learning. ReMix is a general strategy for training deep models on imbalanced data that is simple to implement and computationally efficient. Our empirical results on benchmark tabular and image datasets show that deep learning models training on imbalanced datasets using ReMix achieve equivalent or better predictive performance, as well as better calibration, than models trained with MixUp and traditional resampling and cost-adjustment methods.



\bibliography{library}

\end{document}